# A New Sufficient Condition for 1-Coverage to Imply Connectivity

Seyed Hossein Khasteh

*ACECR*

*Nasir Branch*
*Tehran, Iran*
H_khasteh@ce.sharif.edu

Saeid Bagheri Shouraki

*Department of Electrical Engineering*

*Sharif University of Technology*
*Tehran, Iran*
bagheri-s@sharif.edu

Ali Akbar Kiaei

*ACECR*

*Nasir Branch*
*Tehran, Iran*
kiaei@ce.sharif.edu

**Abstract**

*An effective approach for energy conservation in wireless sensor networks is scheduling sleep intervals for extraneous nodes while the remaining nodes stay active to provide continuous service. For the sensor network to operate successfully the active nodes must maintain both sensing coverage and network connectivity, It proved before if the communication range of nodes is at least twice the sensing range, complete coverage of a convex area implies connectivity among the working set of nodes. In this paper we consider a rectangular region $A = a*b$, such that $R_s \leq a, R_s \leq b$, where $R_s$ is the sensing range of nodes. and put a constraint on minimum allowed distance between nodes($R_s$). according to this constraint we present a new lower bound for communication range relative to sensing range of sensors($\sqrt{2+\sqrt{3}}*R_s$) that complete coverage of considered area implies connectivity among the working set of nodes; also we present a new distribution method, that satisfy our constraint.*

## 1. Introduction

Recent technological advances have led to the emergence of pervasive networks of small, low-power devices that integrate sensors and actuators with limited on-board processing and wireless communication capabilities. These sensor networks open new vistas for many potential applications, such as battlefield surveillance, environment monitoring and biological detection [1], [2], [3].
Sensing is only one responsibility of a sensor network. To operate successfully, a sensor network must also provide satisfactory connectivity so that nodes can communicate for data fusion and reporting to base stations. the relationship between connectivity and coverage depends on the ratio of the communication range to the sensing range. However, it is easily seen that a connected network may not guarantee its coverage regardless of the ranges. This is because coverage is concerned with whether *any* location is uncovered while connectivity only requires all locations of active nodes are connected. Hence we focus on analyzing the condition for a *covered network* to guarantee connectivity.
Wang *et al.* [7], proved that in a convex region if $2R_s \leq R_c$ then 1-Coverage Imply Connectivity, also Zhang *et al.* [8], proved that $2R_s \leq R_c$ is a necessary condition for 1-Coverage to Imply Connectivity.
In the rest of this paper, we first formulate the problem of coverage and connectivity in Section 2; a new sufficient condition for Complete Coverage to Imply Connectivity is presented in Section 3. We present a new distribution method, in Section 4 and conclude the paper in Section 5.

## 2. Problem Formulation

We use the same problem formulation that used in [7], several coverage models [4], [5], [6], have been proposed for different application scenarios. In this paper, we assume a point $p$ is *covered* (monitored) by a node $v$ if their Euclidian distance is less than the sensing range of $v$, $R_s$, i.e., $|pv| < R_s$. We define the *sensing circle* $C(v)$ of node $v$ as the boundary of $v$'s coverage region. We assume that any point $p$ on the sensing circle $C(v)$ (i.e., $|pv| = R_s$) is not covered by $v$. Although this definition has an insignificant practical impact, it simplifies our geometric analysis in following sections. Based on the above coverage model, we define a convex region $A$ (that contains at least one sensing circle) as having a *coverage degree* of $K$ (i.e., being *K-covered*) if every location inside $A$ is covered by at least $K$ nodes. Practically speaking, a network with a higher degree of coverage can achieve higher sensing accuracy and be more robust against sensing failures. The coverage configuration problem can be formulated as follows. Given a convex coverage region $A$, and a coverage degree $K$ specified by the application (either before or after deployment), we must maximize the number of sleeping nodes under the constraint that the remaining nodes must guarantee $A$ is K-covered.

In addition, we assume that any two nodes $u$ and $v$ can directly communicate with each other if their Euclidian distance is less than a communication range $R_c$, i.e., $|uv| < R_c$. Given a coverage region $A$ and a sensor coverage degree $K$, the goal of an integrated coverage and connectivity configuration is maximizing the number of nodes that are scheduled to sleep under the constraints that the remaining nodes must guarantee: 1) $A$ is at least $K$–covered, and 2) all active nodes are connected.

## 3. Sufficient Condition for 1-Coverage to Imply Connectivity

Define the graph $G(V, E)$ to be the communication graph of a set of sensors, where each sensor in the set is represented by a node in $V$, and for any node $x$ and $y$ in $V$, the edge $(x, y)$ exist in $E$ if and only if the Euclidean distance between $x$ and $y$ is less than $R_c$, i.e., $|xy| < R_c$. Node $v$ and $u$ are connected in $G(V, E)$ if and only if a *network path* consisting of consecutive edges in $E$ exists between node $u$ and $v$, we put a constraint on minimum allowed distance between any two nodes, the minimum allowed distance between any two nodes is $R_s$.

Now we assume a scenario like this (Figure 1.): We assume a rectangular region $A = a*b$, such that $R_s \leq a, R_s \leq b$, we assume this region has a complete coverage, for any two nodes $u$ and $v$ in region $A$, let $P_{uv}$ be the line segment joining them. Since region $A$ is convex, $P_{uv}$ remains entirely within $A$. Hence any point on $P_{uv}$ is at least 1-covered. consider the intersection point of $C(u)$ and $P_{uv}$ and name this point as $w$, according to our definition, this point not covered by $u$, We name the node that covers $w$ as $x$, also we name the intersection points of $C(u)$ and $C(x)$ as $s, t$.

**Lemma 1.** $x$ is closer to $v$ than $u$.

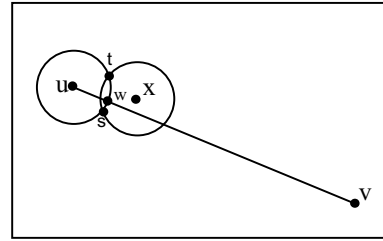

Figure 1, a typical node configuration scenario in a sensor network.

**Proof.** $x$ covers $w$ so the Euclidian distance between $x$ and $w$ is less than $R_s$. $|xw| < R_s = |uw|$. So according to triangle inequality we have:
$|xv| < |xw| + |wv| < |uw| + |wv| = |uv|$, If $x$ lie on the $P_{uv}$ The above inequality is apparent.

**Lemma 2.** If $\sqrt{2+\sqrt{3}} * R_s \leq R_c$ then node $u$ is connected to node $x$ in $G(V, E)$.

**Proof.** (by contradiction) If the Euclidian distance between $x$ and $u$ is less than $R_c$ then the lemma is correct, so we assume that $R_c \leq |ux|$. In this case the distance between point $s$ and $t$ should be less than $R_s$ because otherwise if this distance be at least $R_s$, the angles $\angle u$ in triangle $\triangleright tus$ and $\angle x$ in triangle $\triangleright txs$ are at least $60°$ (sides $|tu|$ and $|su|$ in triangle $\triangleright tus$, and sides $|tx|$ and $|sx|$ in triangle $\triangleright txs$, are all equal to $R_s$; so according to Law of sines in triangles($\frac{a}{\sin A} = \frac{b}{\sin B} = \frac{c}{\sin C}$), angles $\angle u$ in triangle $\triangleright tus$ and $\angle x$ in triangle $\triangleright txs$ must be at least $60°$) so the angle $\angle t$ in triangle $\triangleright utx$ is at most $120°$ (because of symmetry the line $|ux|$ is bisector of angles $\angle u$ in triangle $\triangleright tus$ and $\angle x$ in triangle $\triangleright txs$ so the angles $\angle u$ and $\angle x$ in triangle $\triangleright utx$ are at least $30°$ and so because the sum of three angles of every triangle is $180°$ the angle $\angle t$ in triangle $\triangleright utx$ is at most $120°$) now according to Law of cosines in triangles($c^2 = a^2 + b^2 - 2ab\cos C$), in triangle $\triangleright utx$ we have

$$|ux|^2 = |ut|^2 + |xt|^2 - 2|ut|*|xt|*\cos\angle utx$$

also we know $|ut| = |xt| = R_s$, and, $\angle utx \leq 120$, and cosine is a decreasing function in $[0,180°]$ so we have

$$|ux|^2 = R_s^2 + R_s^2 - 2R_s^2*\cos\angle utx \leq 2R_s^2 - 2R_s^2*\cos 120° = 3R_s^2$$

So we have $|ux| \leq \sqrt{3}*R_s$ and this contradict with our assumption that $\sqrt{2+\sqrt{3}}*R_s \leq R_c$ and $R_c \leq |ux|$. Also we know that $\sqrt{3} < \sqrt{2+\sqrt{3}}$.

So we have $|ts| < R_s$, now we prove that at least one of the points $s$ or $t$ are inside region $A$.

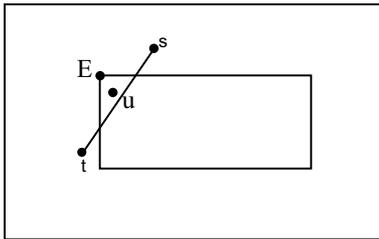

Figure 2, configuration of a line segment that intersect with two joint sides of a rectangular region.

For this, first we assume that both points $s$ and $t$ are outside the region $A$, so because the intersection of $|ts|$ and $|ux|$ is inside the region $A$ (the region $A$ is convex and both points $u$ and $x$ are inside region $A$ and so all of the points of $|ux|$ inside region $A$ )line $|ts|$ intersect with two sides of region $A$, these two side can't be two parallel sides of region $A$ because in this case the length of $|ts|$ should be more than the length of the other sides of region $A$ and this can't be true because we assume that $A = a*b$ is a rectangular region such that $R_s \leq a, R_s \leq b$, also we prove before $|ts| < R_s$. So $|ts|$ intersect with two joint sides of region $A$ (Figure 2.) in this case we have $90° \leq \angle sEt \leq \angle sut$ (points $u$ and $x$ are in two different sides of line $|ts|$ and we assume that $u$ is in the side that restricted by two joint sides of the rectangle and line $|ts|$.) and this can't be true because we prove before $|ts| < R_s$ and that imply $\angle sut \leq 60°$.

So at least one of the points $s$ or $t$ are inside region $A$, we assume that, point $t$ is inside region $A$. So the point $t$ is at least 1-covered, we name the node that covers $t$ as $y$ (Figure 3).

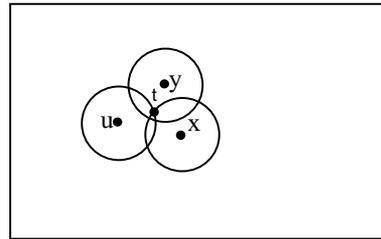

Figure 3, configuration of three nodes that distance between them and a point like $t$ is less than $R_s$.

Now we prove that $|uy| < R_c$ and $|yx| < R_c$, for this first we assume that $R_c \leq |uy|$, in this case the angle $\angle ytu$ must be greater than or equal to $150°$ because we have $|ut| = R_s$ and $|yt| < R_s$ and $R_c \leq |uy|$, now according to law of cosines in triangles($\triangleright uty$), we have

$$|uy|^2 = |ut|^2 + |yt|^2 - 2|ut|*|yt|*\cos\angle ytu$$

Now if the angle $\angle ytu$ is less than $90°$ then it's cosine is positive and we have

$$|uy|^2 < 2R_s^2 \Rightarrow |uy| < \sqrt{2}*R_s < \sqrt{2+\sqrt{3}}*R_s = R_c$$

And if the angle $\angle ytu$ is greater than $90°$ then it's cosine is negative and we have

$|uy|^2 < 2R_s^2 - 2R_s^2*\cos\angle ytu$, if the angle $\angle ytu$ is less than $150°$ since cosine is a decreasing function in $[0,180°]$ we have

$$2R_s^2 - 2R_s^2*\cos\angle ytu < 2R_s^2 - 2R_s^2*\cos 150°$$
$$= (2+\sqrt{3})*R_s^2 \Rightarrow |uy| < \sqrt{2+\sqrt{3}}*R_s = R_c,$$

and this can't be true because we assume that $R_c \leq |uy|$, so the angle $\angle ytu$ is greater than or equal to $150°$, similarly we can prove that angle $\angle utx$ is greater than or equal to $150°$, so the angle $\angle ytx$ is less than or equal to $60°$, also we have $|xt| = R_s$ and $|yt| < R_s$, now according to law of cosines in triangles we have

$$|yx|^2 = |xt|^2 + |yt|^2 - 2|xt|*|yt|*\cos\angle ytx =$$
$$R_s^2 + (|yt| - 2R_s * \cos\angle ytu)*|yt|,$$ since cosine is a decreasing function in $[0, 180°]$ and also $|yt| < R_s$ we have

$$|yx|^2 < R_s^2 + (|yt| - R_s)*|yt| < R_s^2 \Rightarrow |yx| < R_s$$

And this can't be true because we put a constraint on minimum allowed distance between two nodes that must be greater than or equal to $R_s$, so the assumption of $R_c \leq |uy|$ is incorrect and we have $|uy| \leq R_c$, similarly it proved that $|yx| < R_c$. So node $u$ can communicate with node $y$, and node $y$ can communicate with node $x$, so node $u$ is connected to node $x$ in $G(V,E)$.

**Theorem 1.** *For a set of sensors that at least 1-cover a rectangular region $A = a*b$ such that $R_s \leq a, R_s \leq b$ the communication graph is connected if $\sqrt{2+\sqrt{3}} * R_s \leq R_c$.*

**Proof.** For any two nodes $u$ and $v$, according to lemma 1. and lemma 2. we know that $u$ can communicate with a node like $x$ that is closer to $v$ than $u$, the node $x$ is also can communicate with another node that is closer to $v$ than $x$, so we can make a chain that begin by $u$ and every node in this chain is closer to $v$ than previous nodes and can communicate with previous and next nodes in the chain, since the number of nodes and the distance between $u$ and $v$ is finite, this chain can't be infinite and finally reach to $v$, so $u$ and $v$ are connected in communication graph. Since nodes $u$ and $v$ are selected arbitrary the communication graph is connected.

## 4. a New Distribution Method

In this section we present a new distribution method, we assume an arbitrary distribution that at least 1-cover a convex region $A$ and change it such that the distance between any pair of sensors not be less than $R_s$.

**Lemma 3.** *In a circle with radius $R_s$, except the center of circle, there is at most 6 points that distance between any pair of them is at least $R_s$.*

**Proof.** First we assume that there is at least 7 points in a circle with radius $R_s$ that distance between any pair of them is at least $R_s$, now we draw a radius in the circle and sweep the circle in clockwise direction and in order that we meet the points, we number the points.(Figure 4.).

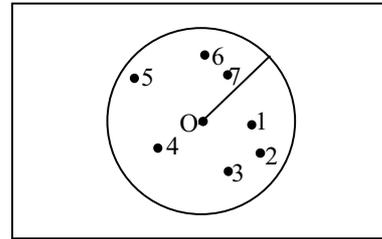

Figure 4, a typical configuration of 7 points in a circle that are numbered in clockwise order.

because we except the center of circle we have 7 triangles: $\triangle 7O1$, $\triangle 1O2$, $\triangle 2O3$, $\triangle 3O4$, $\triangle 4O5$, $\triangle 5O6$ and $\triangle 6O7$, now according to law of sines in triangles, we have that the angle $\angle 7O1$ in triangle $\triangle 7O1$ is at least $60°$, also in triangles $\triangle 1O2$, $\triangle 2O3$, $\triangle 3O4$, $\triangle 4O5$, $\triangle 5O6$ and $\triangle 6O7$ we have that angles $\angle 1O2$, $\angle 2O1$, $\angle 3O4$, $\angle 4O5$, $\angle 5O6$ and $\angle 6O7$ are at least $60°$, and this can't be true because the angle $\angle O$ becomes more than $360°$, so there is at most 6 points in a circle with radius $R_s$ that distance between any pair of them is at least $R_s$.

Now we assume an arbitrary distribution that at least 1-cover a convex region $A$ and change it such that the distance between any pair of sensors not be less than $R_s$, we choose a pair of sensors that the distance between them is less than $R_s$, we name this two sensors $v$ and $u$, we eliminate node $u$, now if there are some points in region $A$ that not covered by any sensor we put a sensor in one of them, after that we calculate the covered region again if still there are some points that are not covered by any sensor we put a sensor in one of that points and repeat this until all the region $A$ be at least 1-covered, we need to put at most 6 new sensors in the region $A$, because we put a sensor in a point that not covered by any other sensor and so it's distance from all other nodes is at least $R_s$, so the distance between any pair of new sensors is at least $R_s$, also all of this sensors be in a circle with radius $R_s$, because before eliminating node $u$, all the region $A$ was at least 1-covered, so the points that become uncovered were covered by node $u$, also none of this new sensor be in the center of circle that cover all of them ($C(u)$) because the distance between node $u$ and node $v$ is less than $R_s$, so the position of node $u$ is covered by node $v$, now according to lemma 3; we need to put at most 6 new sensors. With this method there is no new pair of sensors that the distance between them is less than $R_s$, because the positions of new sensors be uncovered before putting these new sensors and so the distance between the position of these new sensors and any other sensor is at least $R_s$.

By this method we eliminate one pair of sensors that distance between them is less than $R_s$ and because the number of sensors and so the number of these pairs is finite, finally we reach a distribution that the distance between any pair of sensors is at least $R_s$ and the region $A$ is at least 1-covered.

## 5. Conclusions and future works

In this paper we present a tighter sufficient condition for 1-coverage to imply connectivity in wireless sensor networks and corresponding sensor location method, which can facilitate the sensor location problem and save the energy.

We put a constraint on the minimum allowed distance between any two sensors, this constraint is not a restrictive condition because we present a new distribution method for sensors with sensing range $R_s$; which previously cover a convex region $A$; that the distance between any two sensor is at least $R_s$, in this method we put at most 6 new sensors instead of any previous sensor, but in most practical application there is no need to put 6 new sensors, and few sensors are enough. also we consider a rectangular region $A = a*b$, such that $R_s \leq a, R_s \leq b$, all the proofs presented work for a convex polygonal region that any angle of it is at least $60°$, and length of all sides of it are at least $R_s$, almost all the real regions that should be covered satisfy these conditions and so this is not a restrictive constraint.

By this result we can use sensors with shorter communication range, so these sensors because need to send messages to shorter distance consume less energy.

In the future, we will extend our solution to handle other convex regions and other constraints on minimum allowed distance between sensors, and make our condition tighter and save much energy.

## 6. References

[1] D Estrin, R Govindan, J S Heidemann, and S Kumar Next century challenges: *"Scalable coordination in sensor networks"*, In Proc. of ACM MobiCom'99, Washington., 1999.

[2] J M Kahn, R H Katz and K S J Pister, Next century challenges: *"Mobile networking for "smart dust""*, In Proc. of ACM MobiCom'99, 1999.

[3] A Mainwaring, J Polastre, R Szewczyk and D Culler, *"Wireless sensor networks for habitat monitoring"*, In First ACM International Workshop on Wireless Workshop in Wireless Sensor Networks and Applications (WSNA 2002), 2002.

[4] S. Meguerdichian, F. Koushanfar, M. Potkonjak, and M. Srivastava, *"Coverage Problems in Wireless Ad-Hoc Sensor Networks."*, INFOCOM'01, Vol 3, pp. 1380-1387, April 2001.

[5] S. Meguerdichian, F. Koushanfar, G. Qu, and M. Potkonjak, *"Exposure in Wireless Ad Hoc Sensor Networks."*, Procs. Of 7th Annual International Conference on Mobile Computing and Networking (MobiCom'01), pp. 139-150, July 2001.

[6] S. Meguerdichian and M. Potkonjak. *"Low Power 0/1 Coverage and Scheduling Techniques in Sensor Networks."*, UCLA Technical Reports 030001. January 2003.

[7] X Wang, G Xing, Y Zhang, C Lu, R Pless and C Gill *"Integrated coverage and connectivity configuration in wireless sensor networks"*, In *ACM Sensys'03*. 2003.

[8] Honghai Zhang and Jennifer C. *" Maintaining Sensing Coverage and Connectivity in Large Sensor Networks"*, *Ad Hoc & Sensor Wireless Networks,* Vol. 1, pp. 89–124, 2005.